\documentclass[runningheads]{llncs}
\usepackage{graphicx}
\usepackage{amssymb}
\usepackage{subfigure}
\usepackage{array}
\usepackage{multirow}
\usepackage{bbding}
\usepackage{hyperref}

\begin{document}
\title{BESTMVQA: A Benchmark Evaluation System for Medical Visual Question Answering\thanks{This work was done when Xiaojie Hong worked for the project of Meetyou AI Lab.}}
\author{ Xiaojie Hong\inst{1} \and
Zixin Song\inst{1} \and
Liangzhi Li\inst{2 (}\Envelope\inst{)} \and
Xiaoli Wang\inst{1 (}\Envelope\inst{)}\orcidID{0000-0002-8677-9080} \and 
Feiyan Liu\inst{1}}

\authorrunning{X. Hong et al.}
\titlerunning{BESTMVQA: A Benchmark Evaluation System for Med-VQA}
\institute{School of Informatics, Xiamen University, Xiamen, China
\email{\{xjhong,zxsong,feiyanliu\}@stu.xmu.edu.cn, xlwang@xmu.edu.cn}\\
\and
Meetyou AI Lab, Xiamen, China\\ 
\email{liliangzhi@xiaoyouzi.com}}
\maketitle 
\begin{abstract}
Medical Visual Question Answering (Med-VQA) is a very important task in healthcare industry, which answers a natural language question with a medical image. Existing VQA techniques in information systems can be directly applied to solving the task. However, they often suffer from (\textit{i}) the data insufficient problem, which makes it difficult to train the state of the arts (SOTAs) for the domain-specific task, and (\textit{ii}) the reproducibility problem, that many existing models have not been thoroughly evaluated in a unified experimental setup. To address these issues, this paper develops a Benchmark Evaluation SysTem for Medical Visual Question Answering, denoted by BESTMVQA. Given self-collected clinical data, our system provides a useful tool for users to automatically build Med-VQA datasets, which helps overcoming the data insufficient problem. Users also can conveniently select a wide spectrum of SOTA models from our model library to perform a comprehensive empirical study. With simple configurations, our system automatically trains and evaluates the selected models over a benchmark dataset, and reports the comprehensive results for users to develop new techniques or perform medical practice. Limitations of existing work are overcome (\textit{i}) by the data generation tool, which automatically constructs new datasets from unstructured clinical data, and (\textit{ii}) by evaluating SOTAs on benchmark datasets in a unified experimental setup. The demonstration video of our system can be found at \url{https://youtu.be/QkEeFlu1x4A}. Our code and data will be available soon.

\keywords{Medical Visual Question Answering \and Benchmark Evaluation System \and Comprehensive Empirical Study}
\end{abstract}
\section{Introduction}
\begin{figure}[t]
	\centering
	\begin{minipage}{0.475\linewidth}
		\centering
		\includegraphics[width=0.9\linewidth]{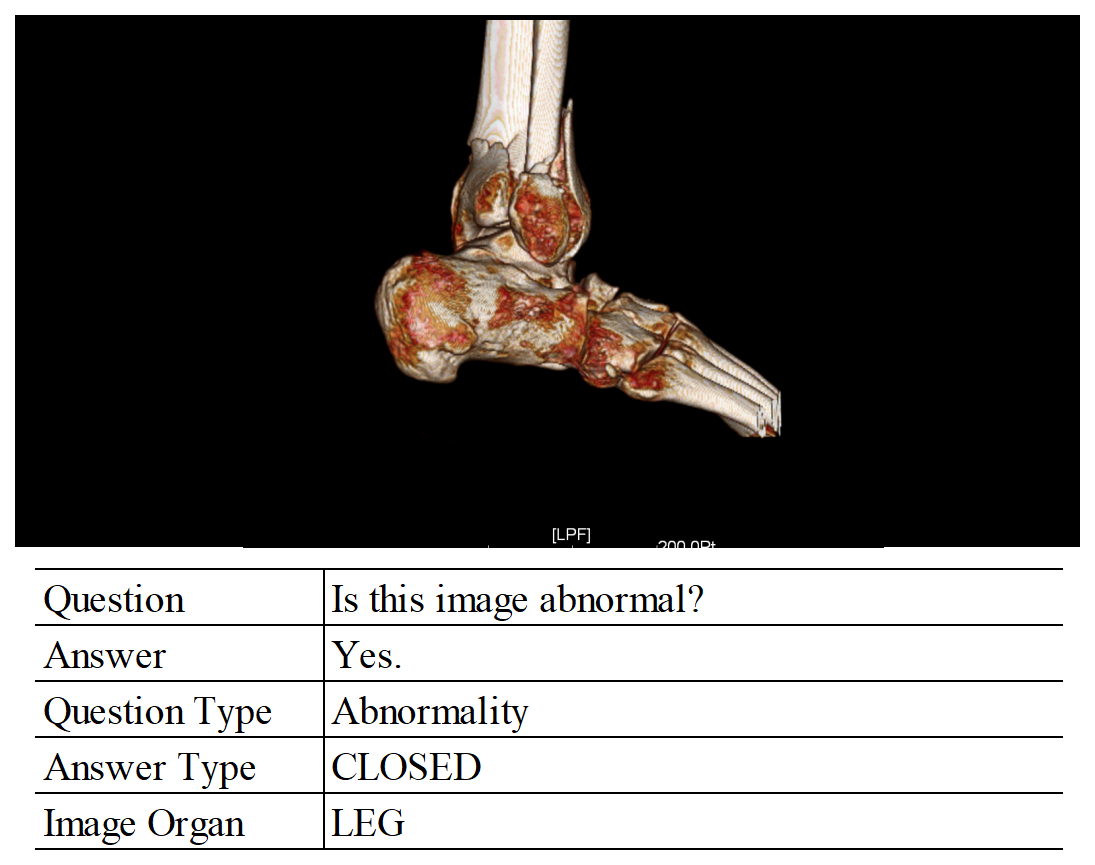}
		\caption{An example of Med-VQA}
		\label{ovqa_data}%文中引用该图片代号
	\end{minipage}
	%\qquad
	\begin{minipage}{0.515\linewidth}
		\centering
		\includegraphics[width=0.9\linewidth]{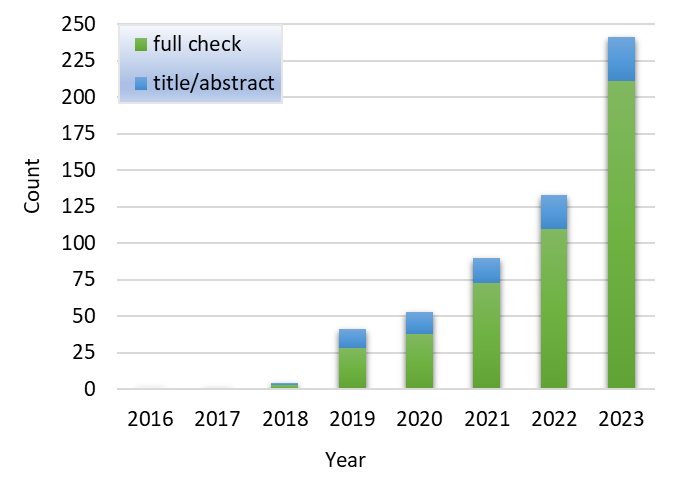}
		\caption{Publications on Med-VQA since 2016}
		\label{paper}%文中引用该图片代号
	\end{minipage}
\end{figure}

Medical visual question answering is a challenging task in healthcare industry, which answers a natural language question with a medical image. Fig.\ref{ovqa_data} shows an example of the Med-VQA data. It may aid doctors in interpreting medical images for diagnoses with responses to close questions, or help patients with urgent needs get timely feedback on open questions \cite{huang2023effective}. It is a challenging problem in computer vision and natural language processing, which processes multi-modal information of visual images and textual language. Different from general VQA, Med-VQA requires substantial prior domain-specific knowledge to thoroughly understand the contents and semantics of medical visual questions.

Many exiting techniques in information systems contribute to solving this task (e.g., \cite{gong2022vqamix}). However, they generally suffer from the data insufficient problem. They need to be trained on well-annotated large-scale datasets, to learn enough domain-specific knowledge for understanding medical visual questions. Several works have focused on constructing Med-VQA datasets \cite{ben2019vqa,he2020pathological,huang2022ovqa,lau2018dataset,liu2021slake}. However, these datasets seem to be a drop in the bucket. Several works employ data augmentation method to tackle the problem. VQAMix \cite{gong2022vqamix} has focused on generating Med-VQA training samples. However, it may incur noisy samples that affect the performance of models. Current work have adopted transfer learning to pre-train a visual encoder on external medical image-text pairs to capture suitable visual representations for subsequent cross-modal reasoning \cite{eslami2021does,gong2022vqamix,huang2023effective}. They have achieved significant success by performing pre-training using large-scale pairs of medical images and text, without additional manual annotations. However, they have not been thoroughly evaluated in benchmark settings.

To address the problems, we develop BESTMVQA, which is a benchmark evaluation system for Med-VQA. We first provide a data generation tool for users to automatically construct new datasets from self-collected clinical data. We implement a wide spectrum of SOTA models for Med-VQA in a model library. Accordingly, users can conveniently select a benchmark dataset and any model in model library for medical practice. Our system can automatically train the models and evaluate them over the selected dataset, and present a final comprehensive report to users. With our system, researchers can comprehensively study SOTA models and their applicability in Med-VQA. The impact of our contributions also can be inferred from Fig.\ref{paper}, which shows the significant increase in Med-VQA publications since 2016. We provide a unified evaluation system for users to (\textit{i}) reveal the applicability of SOTA models to benchmark datasets, (\textit{ii}) conduct a comprehensive study of the available alternatives to develop new Med-VQA techniques, and (\textit{iii}) perform various medical practice.

\section{Research Scope and Task Description}\label{sec:task_des}
The research scope is tailored to two types of readers: (\textit{i}) Researchers who require Med-VQA techniques to perform downstream tasks; (\textit{ii}) Contributors in the research community of Med-VQA who need to thoroughly evaluate the SOTAs.

Medical visual question answering is defined as a domain-specific task that takes a medical image and a clinical question about the image as the input and output an answer in natural language. It generally requires prior domain-specific knowledge to thoroughly understand the contents and semantics of medical visual questions, resulting in additional challenges compared against the general VQA task. The lack of well-annotated large-scale datasets makes it hard to learn enough medical knowledge. To address the challenge, current work typically pre-train a visual encoder on large unlabeled medical image-text pairs.

\begin{figure}
\includegraphics[width=\textwidth]{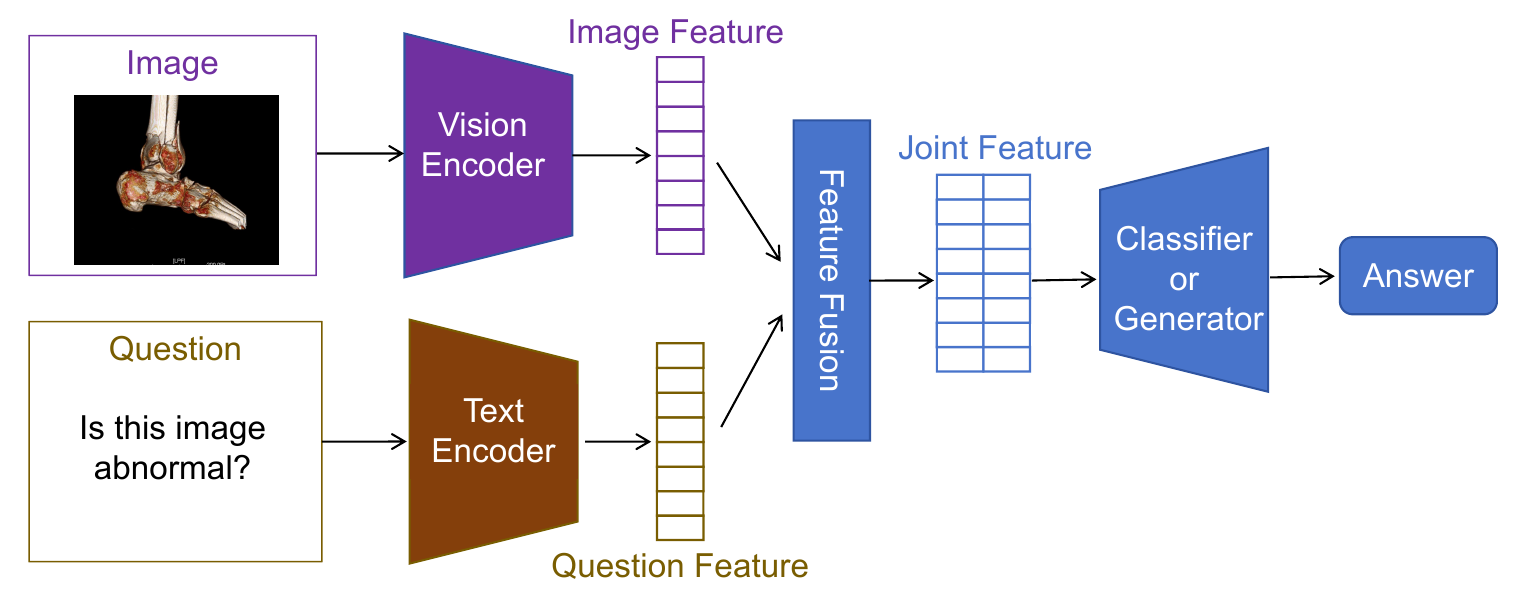}
\caption{The architecture of mainstream Med-VQA models} \label{fig_archi}
\end{figure}

In Fig.~\ref{fig_archi}, Med-VQA models mainly contain four components: vision encoder, text encoder, feature fusion and answer prediction. The vision encoder takes a medical image as the input, then outputs the image features. The text encoder captures the textual features by taking the medical question as the input. The visual and textual features are then fused to generate a joint representation that is finally used as the input of a classifier or generator for predicting the answer.

\begin{table}[t]
\caption{The statistics of considered models, including the parameter size (Params), the training time (Training Time), supporting pre-training or not (Support PT), supporting fine-tuning or not (Support FT) and model category (Model Category). The left value of Training Time represents the smallest training time over all datasets, while the right value is the largest one.}\label{tab1}
    \centering
    \begin{tabular}{|l|l|l|l|l|l|}
    \hline
        Baseline & Params & Training Time & Support PT & Support FT & Model Category \\ \hline
        MEVF \cite{nguyen2019overcoming} & 15M & 0.03h~0.3h & $\times$ & \checkmark & Joint Embedding \\ \hline
        CR \cite{zhan2020medical} & 38M & 0.04h-0.4h & $\times$ & \checkmark & Joint Embedding \\ \hline
        MMQ \cite{do2021multiple} & 20M & 0.5h-3.0h & \checkmark & \checkmark & Joint Embedding \\ \hline
        VQAMix \cite{gong2022vqamix} & 19M & 0.6h-6.0h & $\times$ & \checkmark & Joint Embedding \\ \hline
        CMSA \cite{gong2021cross} & 88M & 1.0h-4.2h & $\times$ & \checkmark & Attention-Based \\ \hline
        MMBERT \cite{khare2021mmbert} & 117M & 1.7h-13.3h & \checkmark & \checkmark & Attention-Based \\ \hline
        PTUnifier \cite{chen2023towards} & 350M & 3.0h-13.0h & \checkmark & \checkmark & Attention-Based \\ \hline
        METER \cite{dou2022empirical} & 320M & 2.5h-18.0h & \checkmark & \checkmark & Attention-Based \\ \hline
        TCL \cite{yang2022vision} & 580M & 1.3h-8.3h & $\times$ & \checkmark & Encoder-Decoder \\ \hline
        MiniGPT-4 \cite{zhu2023minigpt} & 14110M & - & $\times$ & $\times$ & LLMs \\ \hline
    \end{tabular}
\end{table}

\section{Related Work} \label{relatedwork}
Med-VQA is a challenging task that combines natural language processing and computer vision. Early work employing traditional machine learning algorithms suffers from poor performance due to significant differences between visual and textual features \cite{Wu_Teney_Wang_Shen_Dick_Hengel_2016}. Inspired by the success of deep learning in information systems, deep learning models for Med-VQA are reported to have performance gains over traditional models \cite{Srivastava_Murali_Dubey_Mukherjee_2021}. They can be classified into four categories: joint embedding, encoder-decoder, attention-based, and large language models (LLMs). Table~\ref{tab1} shows the statistics of SOTAs we reproduced.

The joint embedding models combine visual and textual embeddings into a final representation. We implement some representative models such as MEVF \cite{nguyen2019overcoming} and CR \cite{zhan2020medical} . MEVF uses MAML \cite{finn2017model} and CDAE \cite{masci2011stacked} to initialize the model weights for visual feature extraction, while CR proposes question-conditioned reasoning and task-conditioned reasoning modules for textual feature extraction. 

For encoder-decoder models, visual and textual features are extracted separately by encoders, and fused in a feature fusion layer. The decoder generates the answer based on the fused features. TCL \cite{yang2022vision} is such a representative model.

The third category employs attention mechanisms to capture representative visual and textual features. We have implemented four representative models, including MMBERT \cite{khare2021mmbert}, CMSA \cite{gong2021cross},  PTUnifier \cite{chen2023towards} and METER \cite{dou2022empirical}. MMBERT \cite{khare2021mmbert} employ Transformer-style architecture to extract visual and textual features. CMSA \cite{gong2021cross} introduce a cross-modal self-attention module to selectively capture the long-range contextual relevance for more effective fusion of visual and textual features.

Recently, LLMs are trained on large amounts of textual data that can help interpret complex and detailed information in medical images. Our model library also provides the MiniGPT-4 \cite{zhu2023minigpt} for generating the linguistic representation of the question in Med-VQA, and supporting further medical practice.

\begin{figure}[t]
\centering
\includegraphics[width=\textwidth]{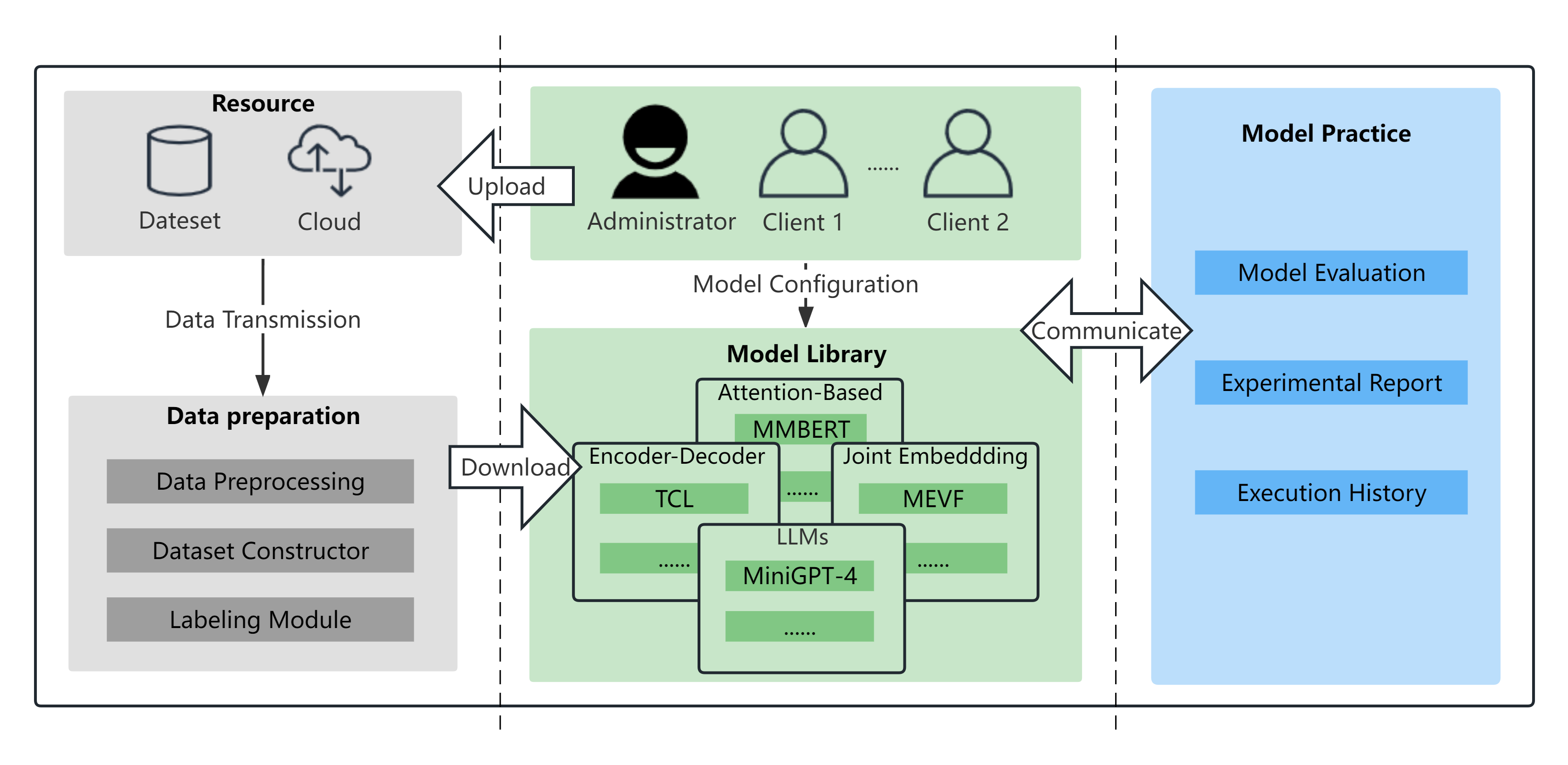} 
\caption{System architecture of our BESTMVQA}
\label{figsys}
\end{figure}

\section{System Overview}
In Fig.\ref{figsys}, our BESTMVQA system has three components: data preparation, model library, and model practice. The data preparation component is developed based on a semi-automatic data generation tool. Users first upload self-collected clinical data. Then, medical images and relevant texts are extracted for medical concept discovery. We provide a human-in-the-loop framework to analyze and annotate medical concepts. To facilitate the effort, we first auto-label the medical concepts by employing the BioLinkBERT-BiLSTM-CRF \cite{yasunaga-etal-2022-linkbert}. Then, professionals can conveniently verify the medical concepts. After that, medical images, medical concepts and diagnosis texts are fed into a pre-trained language model for generating high-quality QA pairs. We employ a large-scale medical multi-modal corpus to pre-train and fine-tune an effective model, which can be easily incorporated into existing neural models for generating medical VQA pairs. our system provides a model library, to avoid duplication of efforts on implementing SOTAs for experimental evaluation. A wide spectrum of SOTAs have been implemented. The detailed statistics of the models can be seen in Section \ref{relatedwork}. Based our library, users can conveniently select a benchmark dataset and any number of SOTAs from our model library. Then, our system automatically performs extensive experiments to evaluate SOTAs over the benchmark dataset, and presents the final report to the user. From our report, the user can comprehensively study SOTAs and their applicability to Med-VQA. Users can also download the experimental reports and the source codes for further practice.

\section{Empirical Study}
Users can conveniently use our BESTMVQA system to systematically evaluate SOTAs on benchmark datasets for Med-VQA. To comprehensively evaluate the effectiveness of the models, we employ the metric of $accuracy$ for open-ended questions, closed-ended questions, and the overall questions. Five datasets are provided for users for model practice to investigate the applicability of models to diverse application scenarios.

\begin{table}[t]
\caption{The statistics of datasets. Here, NI, NQ, and NA represent the number of images, questions and answers, respectively. MeanQL and MeanAL represent the length of questions and answers, respectively.}\label{tab2}
    \centering
    \begin{tabular}{|l|l|l|l|l|l|}
    \hline
        Dataset & NI & NQ & MeanQL & MeanAL & NA \\ \hline
        VQA-RAD \cite{lau2018dataset}  & 314 & 3515 & 6.49 & 1.61 & 557 \\ \hline
        MedVQA-2019 \cite{ben2019vqa} & 4200 & 15292 & 6.88 & 2.12 & 1749 \\ \hline
        SLAKE-EN \cite{liu2021slake} & 642 & 7033 & 8.03 & 1.4 & 234 \\ \hline
        PathVQA \cite{he2020pathological} & 4289 & 32795 & 6.33 & 1.79 & 4946 \\ \hline
        OVQA \cite{huang2022ovqa} & 2000 & 19020 & 8.73 & 3.32 & 1065 \\ \hline
    \end{tabular}
\end{table}

\subsection{Considered Models}
We emphasize the utilization of ``out-of-the-box'' models, defining a model as ``usable out of the box'' if it meets the following criteria: (\textit{i}) publicly available executable source code, (\textit{ii}) well-defined default hyperparameters, (\textit{iii}) no mandatory hyperparameter optimization, and (\textit{iv}) absence of requirements for language model retraining and vocabulary adaptation. To ensure consistent evaluation and practical applicability, all models are expected to generate predictions in a standard format. Adhering to the criteria is essential for models that can help guarantee aligning with the concept of ``out of the box''.

According to this definition, ten models are identified and classified, as shown in Table~\ref{tab1}. These models contain: (\textit{i}) those specifically tailored for Med-VQA, and (\textit{ii}) the application of general VQA models to the medical domain.

\begin{figure}[ht]
		\centering
		\subfigure[OVQA]{
			\includegraphics[width=0.31\textwidth]{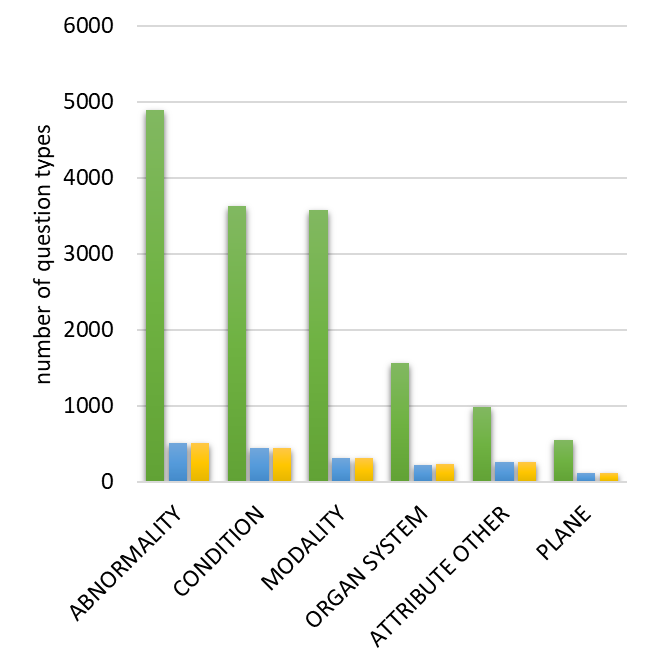}
			\label{fig:image1}
		}
		\subfigure[MedVQA-2019]{
			\includegraphics[width=0.31\textwidth]{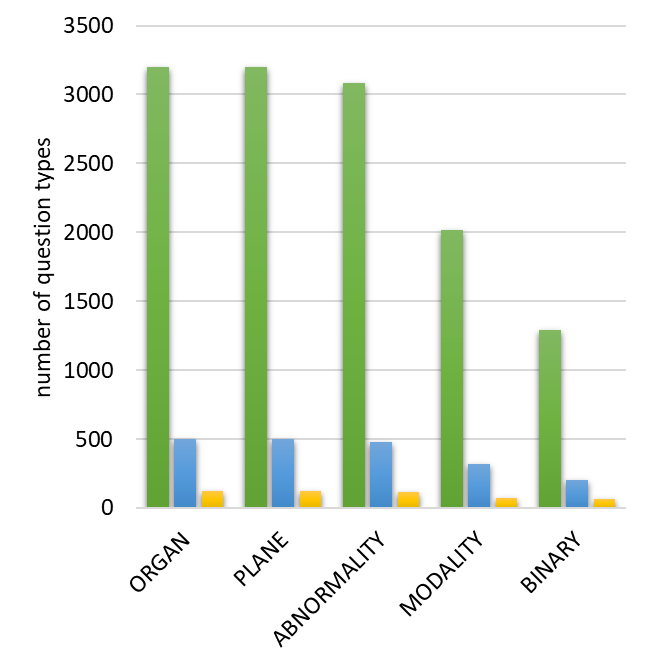}
			\label{fig:image2}
		}
		\subfigure[SLAKE-EN]{
			\includegraphics[width=0.31\textwidth]{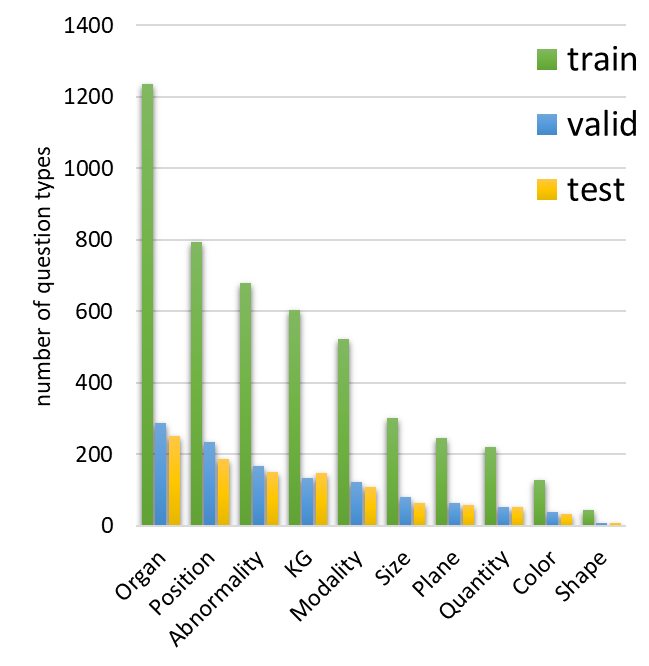}
			\label{fig:image3}
		}
		\\
		\subfigure[PathVQA]{
			\includegraphics[width=0.3\textwidth]{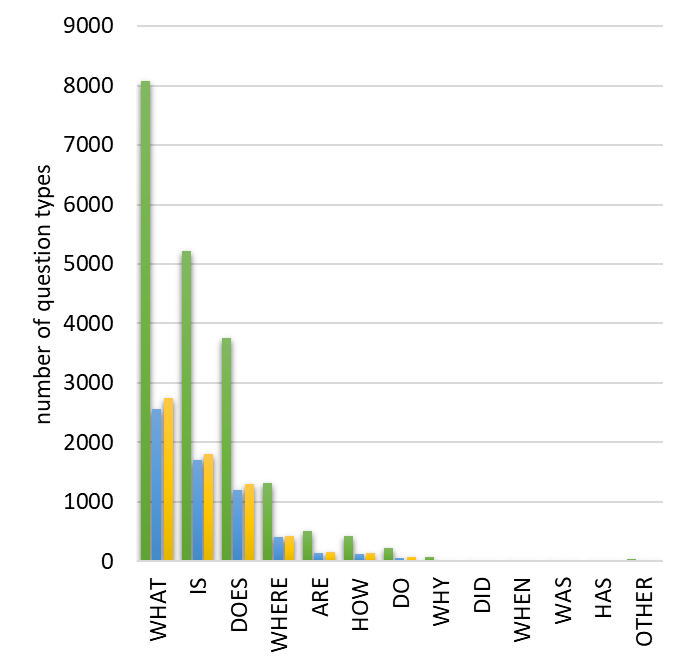}
			\label{fig:image4}
		}
		\subfigure[VQA-RAD]{
			\includegraphics[width=0.3\textwidth]{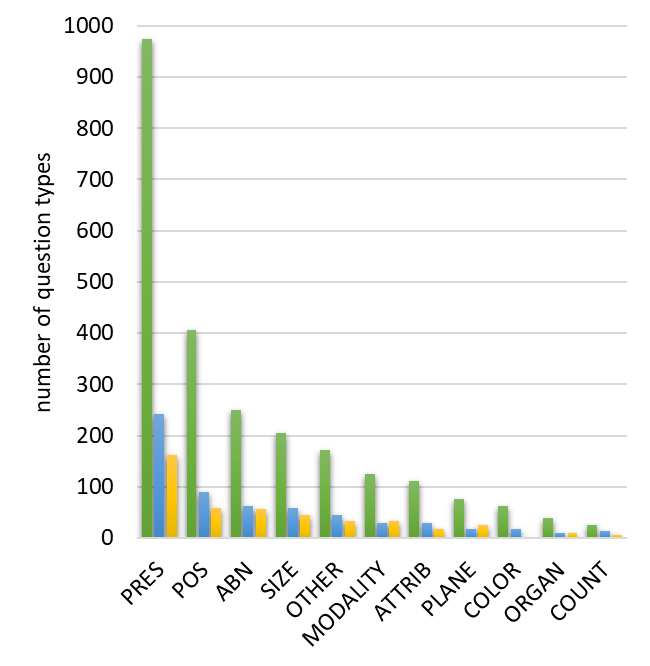}
			\label{fig:image5}
		}
		\caption{Distribution of question types per dataset}
		\label{fig02}
\end{figure}

\begin{table}
\caption{Default values for Batch Size, Learning Rate, and Epoch for each model}\label{tab_para}
    \centering
    \begin{tabular}{|l|l|l|l|}
    \hline
        {\hspace{5pt}}Baseline & Batch Size & Learning Rate & Epoch \\ \hline
        MEVF+SAN & 16 & 1.00E-03 & 20 \\ \hline
        MEVF+BAN & 8 & 1.00E-03 & 20 \\ \hline
        CR & 64 & 1.00E-03 & 40 \\ \hline
        MMQ & 64 & 1.00E-03 & 60 \\ \hline
        VQAMix+SAN & 8 & 1.00E-03 & 80 \\ \hline
        VQAMix+BAN & 8 & 1.00E-03 & 80 \\ \hline
        CMSA & 32 & 1.00E-03 & 60 \\ \hline
        MMBERT & 16 & 1.00E-03 & 80 \\ \hline
        PTUnifier & 8 & 1.00E-05 & 50 \\ \hline
        METER & 32 & 1.00E-05 & 25 \\ \hline
        TCL & 4 & 2.00E-05 & 20 \\ \hline
    \end{tabular}
\end{table}

\subsection{Experimental Setup}
\paragraph{Datasets.} 
All models are evaluated using the following five datasets:  

\textbf{OVQA} \cite{huang2022ovqa} has 2,001 images and 19,020 QA pairs, with each image linked to multiple QA pairs. 

\textbf{VQA-RAD} \cite{lau2018dataset} includes 314 images and 3,515 questions answered by clinical doctors, with 10 question types across the head, chest and abdomen. 

\textbf{SLAKE} \cite{liu2021slake} is a bilingual dataset annotated by experienced doctors, which is represented as SLAKE-EN in English.

\textbf{MedVQA-2019} \cite{ben2019vqa} is a radiology dataset from the ImageClef challenge, which includes 642 images with over 7,000 QA pairs. 

\textbf{PathVQA} \cite{he2020pathological} consists of 32,795 pairs generated from pathological images.

Datasets were chosen for their diversity in sample sizes (refer to Table~\ref{tab2}). For VQA-RAD and SLAKE, we have reorganized the datasets  in a 70\%-15\%-15\% ratio due to the lack of validation sets. As for the other datasets, We use the proportion of the corresponding dataset for split. The detailed statistics for data splits are shown in Table~\ref{tab3}. The distribution of question types is illustrated in Fig.~\ref{fig02}.

\begin{table}[h]
\caption{The statistics of data splits. NI represents the number of images. MaxQL, MinQL and MeanQL represent the max, min and mean length of question, respectively;  NCF and NOF represent the number of close-ended and open-ended questions in the corresponding samples. Note that MedVQA-2019 is not strictly divided into open-ended and closed-ended questions.}\label{tab3}
    \centering
    \begin{tabular}{|l|l|l|l|l|l|l|l|l|}
    \hline
        Dataset & Sample & NI & MaxQL & MinQL & MeanQL & Vocabulary & NCF & NOF \\ \hline
        VQA-RAD (train) & 2451 & 314 & 21 & 3 & 6.43 & 1114 & 1443 & 1008 \\ 
        VQA-RAD (valid) & 613 & 258 & 19 & 3 & 6.42 & 625 & 380 & 233 \\ 
        VQA-RAD (test) & 451 & 203 & 22 & 3 & 6.89 & 538 & 272 & 179 \\ 
        Total & 3515 & 314 & 22 & 3 & 6.49 & 1288 & 2095 & 1420 \\ \hline
        MedVQA-2019 (train) & 12792 & 3200 & 11 & 4 & 6.88 & 98 & - & - \\ 
        MedVQA-2019 (valid) & 2000 & 500 & 11 & 4 & 6.86 & 94 & - & - \\ 
        MedVQA-2019 (test) & 500 & 500 & 11 & 4 & 6.86 & 93 & - & - \\ 
        Total & 15292 & 4200 & 11 & 4 & 6.88 & 98 & - & - \\ \hline
        SLAKE-EN (train) & 4777 & 546 & 21 & 4 & 7.98 & 301 & 1905 & 2872 \\ 
        SLAKE-EN (valid) & 1195 & 484 & 18 & 4 & 8.12 & 265 & 460 & 735 \\ 
        SLAKE-EN (test) & 1061 & 96 & 21 & 4 & 8.11 & 265 & 416 & 645 \\ 
        Total & 7033 & 642 & 21 & 4 & 8.03 & 306 & 2781 & 4252 \\ \hline
        PathVQA (train) & 19755 & 2599 & 37 & 2 & 6.35 & 4161 & 9868 & 9887 \\ 
        PathVQA (valid) & 6279 & 832 & 37 & 2 & 6.24 & 2537 & 3156 & 3123 \\ 
        PathVQA (test) & 6761 & 858 & 42 & 2 & 6.33 & 2608 & 3409 & 3352 \\ 
        Total & 32795 & 4289 & 42 & 2 & 6.33 & 5095 & 16433 & 16362 \\ \hline
        OVQA (train) & 15216 & 2000 & 95 & 4 & 8.63 & 958 & 8037 & 7179 \\ 
        OVQA (valid) & 1902 & 1235 & 62 & 4 & 9.04 & 613 & 830 & 1072 \\ 
        OVQA (test) & 1902 & 1234 & 67 & 4 & 9.26 & 533 & 832 & 1070 \\ 
        Total & 19020 & 2000 & 95 & 4 & 8.73 & 1005 & 9699 & 9321 \\ \hline
    \end{tabular}
\end{table}

\begin{table}
\caption{Experimental results for each baseline on the test set of VQA-RAD, SLAKE-EN, PathVQA, and OVQA datasets. This includes the values of three indicators: Closed-ended Questions $Accuracy$, Open-ended Questions $Accuracy$, and Overall $Accuracy$. The definition of evaluation indicators can be found in Section \ref{metrics}}\label{tab5_1}
    \centering
    \begin{tabular}{|c|c|c|c|c|}
    \hline
         {\hspace{5pt}}Dataset{\hspace{5pt}} &  {\hspace{5pt}}Baseline{\hspace{5pt}} &  {\hspace{5pt}}Closed-ended{\hspace{5pt}} &  {\hspace{5pt}}Open-ended{\hspace{5pt}} &  {\hspace{14pt}}Overall{\hspace{14pt}} \\ \hline
        ~ & MEVF+SAN & 75.4 & 40.2 & 61.4 \\ 
        ~ & MEVF+BAN & 78.3 & 52.5 & 68.1 \\ 
        ~ & CR & 77.2 & 57.6 & 69.4 \\ 
        ~ & MMQ & 75.7 & 56.9 & 68.2 \\ 
        ~ & VQAMix+SAN & 79.4 & 57 & 70.5 \\ 
        VQA-RAD & VQAMix+BAN & 80.9 & 57.5 & 71.6 \\
        ~ & CMSA & 78.5 & 63.7 & 72.5 \\ 
        ~ & MMBERT & 74.3 & 46.9 & 63.4 \\ 
        ~ & PTUnifier & \textbf{86.4} & \textbf{68.2} & \textbf{79.2} \\ 
        ~ & METER & 78.3 & 57 & 69.8 \\ 
        ~ & TCL & 73.5 & 56.4 & 66.7 \\
        ~ & MiniGPT-4 & 27.9 & 28.5 & 28.2 \\ \hline
        ~ & MEVF+SAN & 78.4 & 75.3 & 76.5 \\ 
        ~ & MEVF+BAN & 81 & 75.7 & 77.8 \\ 
        ~ & CR & 76.9 & 78.4 & 77.5 \\  
        ~ & MMQ & 78.4 & 76.7 & 77.4 \\ 
        ~ & VQAMix+SAN & 77.9 & 77.7 & 77.8 \\ 
        SLAKE-EN & VQAMix+BAN & 83.2 & 78.1 & 80.1 \\ 
        ~ & CMSA & 68.3 & 49.1 & 56.6 \\ 
        ~ & MMBERT & 43.3 & 1.9 & 18.1 \\
        ~ & PTUnifier & \textbf{89.4} & \textbf{81.6} & \textbf{84.6} \\ 
        ~ & METER & 87.3 & 79.2 & 82.4 \\ 
        ~ & TCL & 87.5 & 78.4 & 82 \\ 
        ~ & MiniGPT-4 & 25.6 & 27.3 & 26.8 \\ \hline
        ~ &MEVF+SAN & 83.4 & 13.1 & 48.5 \\ 
        ~ & MEVF+BAN & 83.8 & 16.4 & 50.3 \\ 
        ~ & CR & 84.9 & 15.9 & 50.5 \\ 
        ~ & MMQ & 83.2 & 14.3 & 48.9 \\ 
        ~ & VQAMix+SAN & 83.9 & 9.6 & 46.9 \\ 
        PathVQA & VQAMix+BAN & 84.3 & 12.7 & 48.6 \\ 
        ~ & CMSA & 83.7 & 16.1 & 50.2 \\ 
        ~ & MMBERT & 83.2 & 13 & 48.1 \\ 
        ~ & PTUnifier & 85.5 & 10.1 & 48.1 \\ 
        ~ & METER & \textbf{89.9} & 29.8 & 60 \\ 
        ~ & TCL & 88.1 & \textbf{36.9} & \textbf{62.7} \\ 
        ~ & MiniGPT-4 & 11.5 & 13.9 & 12.7 \\ \hline
        ~ &MEVF+SAN & 74.2 & 52.3 & 61.9 \\ 
        ~ & MEVF+BAN & 76.6 & 50.5 & 61.9 \\ 
        ~ & CR & 76.6 & 36.9 & 54.3 \\ 
        ~ & MMQ & 79 & 53.2 & 64.5 \\
        ~ & VQAMix+SAN & 77.6 & 59.1 & 67.2 \\ 
        OVQA & VQAMix+BAN & 79.3 & 57 & 66.8 \\  
        ~ & CMSA & 79.7 & 45.6 & 60.5 \\ 
        ~ & MMBERT & 80.5 & 48.7 & 62.6 \\
        ~ & PTUnifier & \textbf{84.9} & \textbf{60.5} & \textbf{71.3} \\ 
        ~ & METER & 82.1 & 51.7 & 65.1 \\ 
        ~ & TCL & 82.6 & 60.4 & 70.1 \\ 
        ~ & MiniGPT-4 & 37.3 & 46.7 & 42.6 \\ \hline
    \end{tabular}
\end{table}

\begin{table}
\caption{Experimental results for each baseline on the test set of MedVQA-2019. Due to the fact that the MedVQA-2019 is not strictly divided into open-ended and closed-ended question types, the table only contains the values of Overall $Accuracy$.}\label{tab5_2}
    \centering
    \begin{tabular}{|c|c|c|}
    \hline
        {\hspace{5pt}}Dataset{\hspace{5pt}} & {\hspace{5pt}}Baseline{\hspace{5pt}} & {\hspace{14pt}}Overall{\hspace{14pt}} \\ \hline
        ~ &MEVF+SAN & 50 \\ 
        ~ & MEVF+BAN & 47.4 \\ 
        ~ & CR & 46.8 \\ 
        ~ & MMQ & 50 \\
        ~ & VQAMix+SAN & 47.2 \\ 
        MedVQA-2019 & VQAMix+BAN & 49 \\  
        ~ & CMSA & 47.4 \\ 
        ~ & MMBERT & 51.2 \\
        ~ & PTUnifier & 60.3 \\ 
        ~ & METER & \textbf{73.9} \\ 
        ~ & TCL & 63 \\
        ~ & MiniGPT-4 & 14.6 \\ \hline
    \end{tabular}
\end{table}

\paragraph{Implementation details.} 
For dataset processing and pre-training, we use a large-scale publicly available medical dataset called by ROCO \cite{pelka2018radiology}. It contains image-text pairs collected from PubMed\footnote{https://pubmed.ncbi.nlm.nih.gov/}, covering various images such as X-rays, MRI, angiography, etc. It also covers a wide range of body regions, such as the head, neck, teeth, etc. We selected 87,952 non composite radiographic images with relevant captions. For fine-tuning, we follow the training, validation, and testing data splitting according to Table~\ref{tab3}. Five benchmark Med-VQA datasets were used to train and evaluate SOTAs. For fair comparison, we use the same data splits and follow the previous work to partition the datasets. Medical visual questions are usually divided into two categories: closed-ended and open-ended questions. Closed-ended questions are usually answered with ``yes/no'' or other limited options. Open-ended questions have no restrictive structure and can have multiple correct answers. All models are trained on dual graphics NVIDIA RTX V100 GPU. We use the AdamW optimizer with the same preheating steps. See Table~\ref{tab_para} for detailed parameter settings of the model.

\subsection{Evaluation Metrics}\label{metrics}
In order to quantitatively measure the performance of models, we use the $accuracy$ as an evaluation metric, and compute it for two types of questions (open-ended and closed-ended). Let $P_i$ and $L_i$ respectively represent the prediction and ground-truth label of sample $i$ in the test set, and $T$ represents the test set. The $accuracy$ is calculated as follows:

\begin{equation}
accuracy = \frac{1}{|T|} \sum_{i\in T}l(P_i=L_i)
\end{equation}

The generative model such as MiniGPT-4 uses the $n$-grams method to calculate the $accuracy$ of predictions and ground-truths. If it is higher than the manually set threshold $\alpha$, the prediction is judged to be correct.

\subsection{Results}
Table~\ref{tab5_1} and ~\ref{tab5_2} show the $accuracy$ achieved by all the considered models. We can obtain the following observations:

%只是解释了ptunifier为什么在医学领域的数据集上表现更好，是因为他是医学领域的模型。但是没有比较为什么他比其他医学领域的模型表现更好，需要加入他表现好的理由(已解决)
(\textit{i}) Among all baselines, the PTUnifier which is pre-trained in the medical domain performs the best on VQA-RAD, SLAKE-EN and OVQA, but not so well on PathVQA and MedVQA-2019. As for the pre-trained models in general domain, TCL and METER achieve better performance on PathVQA and MedVQA-2019. The possible reason is that PathVQA is collected from a wide range of sources, including textbooks and literature, while MedVQA-2019 is artificially generated and cannot represent formal clinical data. PTUnifier adopts a visual language pre-training framework and unifies the fused encoder and dual encoder, thereby excelling on multi-modal tasks.

%可以结合模型的分类再加一些解释：相比于其他类型的模型，大模型现在的性能还是比较差（待定）
%第二点参考introduction第二点的内容
(\textit{ii}) MiniGPT-4 shows the worst performance on every dataset. Although utilizing massive amounts of data for training, it is still unable to effectively mine the domain-specific knowledge to answer a medical question, resulting in poor performance. In addition, the usage of unappropriate prompts may further degrade the model performance.

(\textit{iii}) Discriminative models (all models except of MiniGPT-4), are more applicable to Med-VQA than generative models (MiniGPT-4). The discriminative models define the Med-VQA task as a classification problem, while generative models focus on simulating and generating data, requiring broader language understanding and visual information processing capabilities. Compared with generative models, discriminative models generally have fewer parameters.

\begin{figure}[ht]
		\centering
		\subfigure[Batch Size]{
			\includegraphics[width=0.475\textwidth]{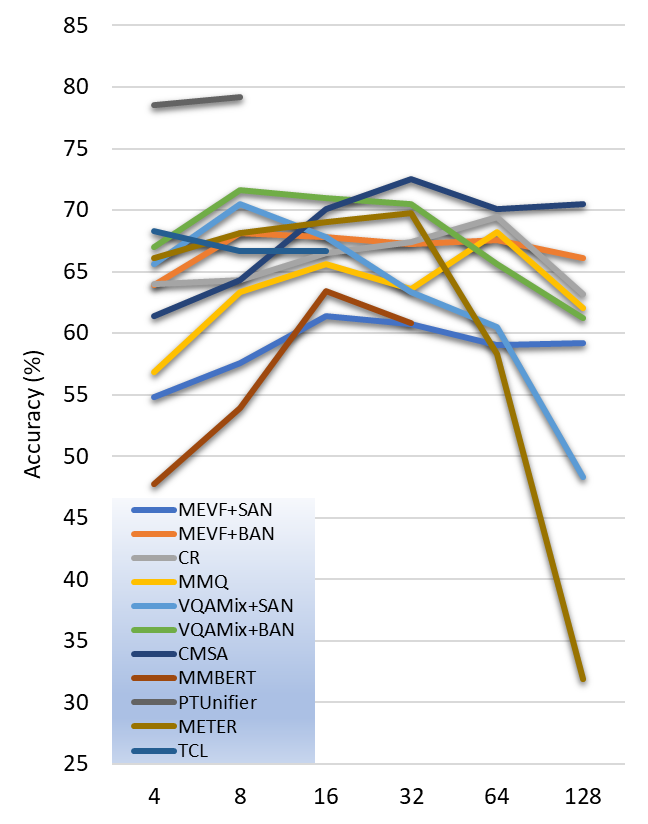}
			\label{fig:bs}
		}
		\subfigure[Learning Rate]{
			\includegraphics[width=0.475\textwidth]{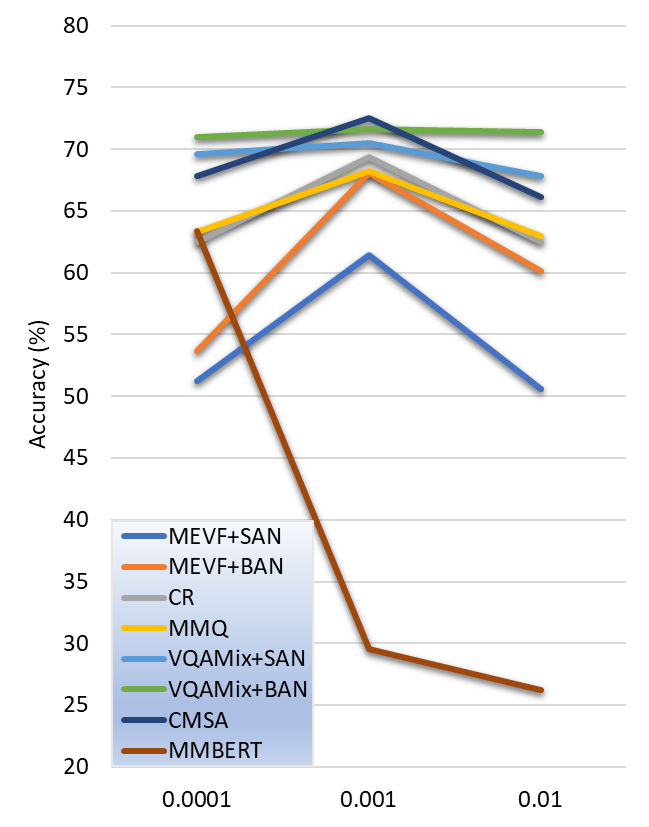}
			\label{fig:lr}
		}
		\caption{Model performance varies with batch size and learning rate}
		\label{fig04}
\end{figure}

(\textit{iv}) The performance of lightweight models such as MEVF, CR, MMQ, and CMSA is significantly inferior to complex models like PTUnifier, TCL, and METER. This is because models like PTUnifier have more parameters and adopt a deeper neural network structure, which is beneficial for learning the alignment between images and texts.
Fig.~\ref{fig04} shows that the values of hyperparameters are determined based on the values set with the best performance on the validation set. The results of each model are obtained by challenging the Batch Size (BZ) and Learning Rate (LR). Due to limited computing power, we only show parts of results: (\textit{i}) The results of MiniGPT-4 are eliminated as it cannot be re-trained and fine-tuned; (\textit{ii}) We show part of results for PTUnifier in Fig.~\ref{fig:bs}, as it requires more computing power for larger values of BZ; (\textit{iii}) Similarly, we show part of results for PTUnifier, TCL, and METER with larger number of parameters in Fig.~\ref{fig:lr}, as the value range of LR is not comparable to that of other models.
\subsection{Detailed Analysis}
In Fig.~\ref{fig:bs}, the performance of each model gradually increases with the increase of BZ values, and then decrease after reaching a saddle point, due to the gradient calculation. However, when BZ is set to a large value, some models converge to local stationary points, such as METER and VQAMix-SAN. In Fig.~\ref{fig:lr}, (\textit{i}) with the increase of LR values, the performance of MMBERT shows a significant decline, and (\textit{ii}) the performance of MEVF, CR, and CMSA first increase and then decrease with the increase of LR values.

\begin{figure}[t]
\centerline{\includegraphics[width=0.95\textwidth]{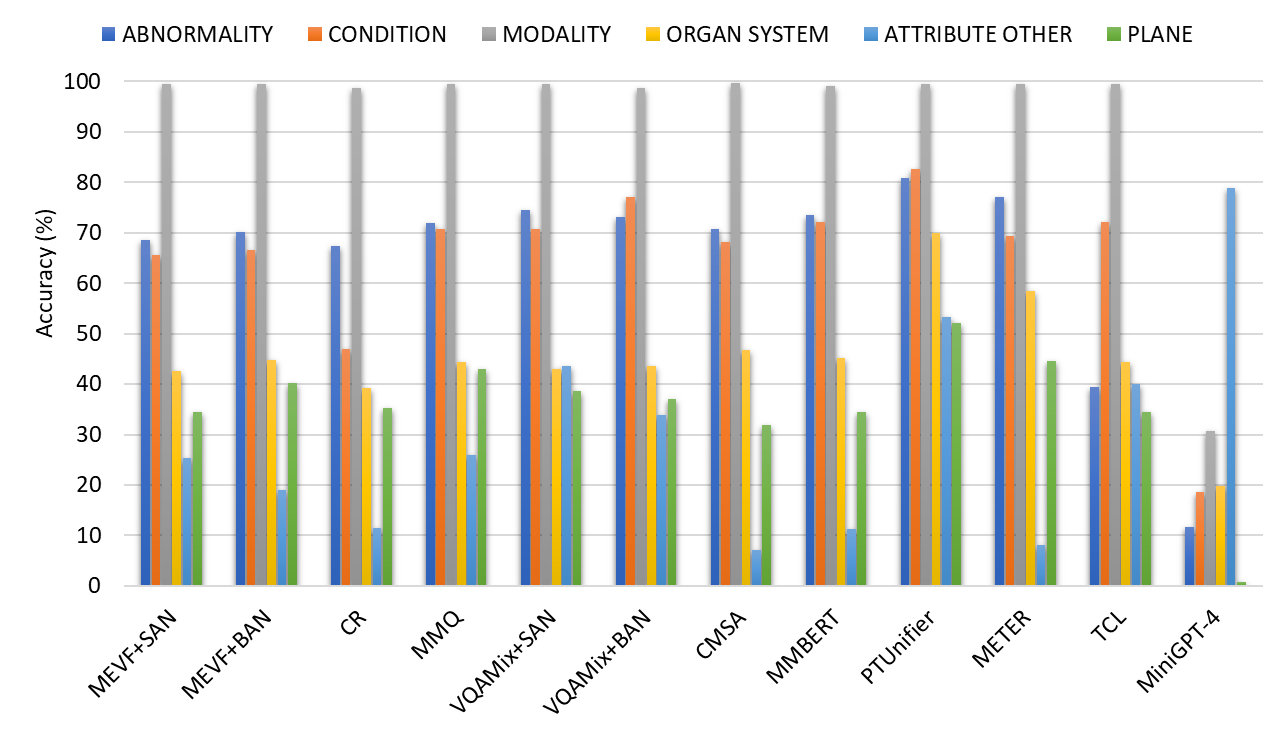}}
\caption{The $Accuracy$ values of different question types for each model in OVQA} \label{fig_ques_type}
\end{figure}

Fig.~\ref{fig_ques_type} shows the results on various question types for each model over the OVQA dataset.

(\textit{i}) All models perform well on the Modality type of questions because MRI or CT image features are obvious, enabling the image encoder to effectively extract image features.

(\textit{ii}) Compared to discriminative models, MiniGPT-4 performs worse on the Plane, Abnormality, and Condition type of questions, as these types of questions are more suitable to the label classification tasks.

(\textit{iii}) MiniGPT-4 achieves significantly better performance than other models on the Attribute Other type of questions. This is because LLMs possess extensive knowledge, providing more comprehensive and complete information when answering descriptive questions.

(\textit{iv}) Among discriminative models, TCL demonstrates better performance on the Attribute Other type of questions. TCL employs a contrastive learning approach in cross-modal and intra-modal self-supervision, which enhances the image-text alignment in the fusion encoder.

(\textit{v}) VQAMix performs well on most types of questions because it incorporates a conditional label combination strategy for data augmentation, allowing for extracting more comprehensive image features.
% (i) Except for minipt4, all other models achieved the best performance on Modality type problems. This is because modularity mainly focuses on image classification problems such as "Was this image taken with an MRI or CT scanner?" For discriminative models such as mevf and cr, using medical VQA as a label classification problem is very helpful, while minipt4 cannot perform well in the same way.

% (ii) Regarding the descriptive problem of Attribute Other, the generative large language model minigpt4 performs the best, significantly outperforming other models and possessing a vast amount of knowledge. Therefore, it can provide more comprehensive and helpful information in solving descriptive problems.

% (iii) TCL excels in solving descriptive problems such as Attribute Other, just like no generative large language model, but is superior to other discriminative models. This is because the model utilizes multiple contrastive learning in cross modal and intra modal self supervision, making it easier for fusion encoders to learn joint multimodal embeddings, which greatly facilitates the resolution of problems such as image text alignment.

% (iv) VQAMix ranks among the top performers in almost all problem types, due to its introduction of Conditional Mixed Labeling (LCL) strategy for data augmentation, which involves linearly stacking samples of the same problem type to extract image features more fully.

\begin{figure}[t]
\includegraphics[width=\textwidth]{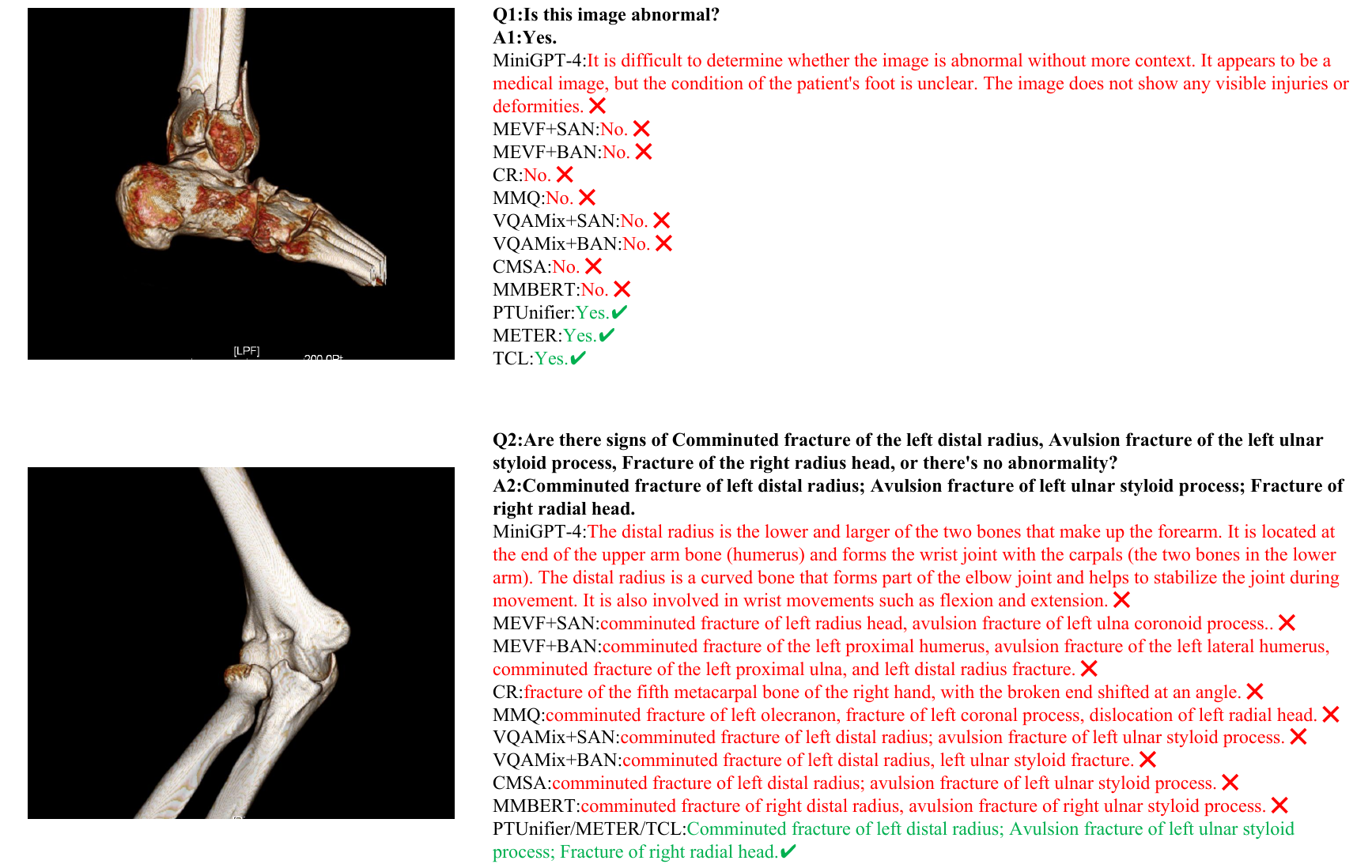}
\caption{Two testing examples selected from OVQA} \label{fig_case}
\end{figure}

\subsection{Qualitative Analysis}
We provide a qualitative comparison of all models. Two examples from the OVQA dataset in Fig.~\ref{fig_case} show that early discriminative models such as MEVF, CR, MMBERT, CMSA, and VQAMix, fail to answer Med-VQA questions, compared to the latest discriminative models such as TCL, METER, and PTUnifier. For example, in the second left figure in Fig.~\ref{fig_case}, the Red Cross indicates that the prediction is wrong, and the green check indicates that the prediction is correct. We observed that although the given question is to consult the abnormal position of orthopedic images, the predicted result position of traditional models such as MEVF is wrong. In the same image, TCL, and other advanced general fields and VQA in the medical field will locate the abnormality to the correct position. This also shows that the advanced VQA deep learning model with large parameters can not only correctly capture the image content as a whole, but also understand the region of interest related to the question, so as to provide the correct answer.

\section{Conclusion}
Deep learning models face additional challenges for Med-VQA, compared with general VQA. It is a urgent need for researchers to perform a comprehensive empirical study on SOTAs over benchmark datasets, to develop new Med-VQA techniques or perform medical practice. Therefore, we implement a benchmark evaluation system for users to support this need. Our system thoroughly compares the user-selected models and reports the comprehensive results obtained from the experiments. Users also can download the datasets, evaluation reports and source code of SOTAs for further practice. In a word, we provide a unified benchmark evaluation system for users to conveniently perform diverse medical practice. The demonstration video of our system can be found at \url{https://youtu.be/QkEeFlu1x4A}.

\bibliographystyle{splncs04}
\bibliography{re}

\begin{thebibliography}{10}
\providecommand{\url}[1]{\texttt{#1}}
\providecommand{\urlprefix}{URL }
\providecommand{\doi}[1]{https://doi.org/#1}

\bibitem{ben2019vqa}
Ben~Abacha, A., Hasan, S.A., Datla, V.V., Demner-Fushman, D., M{\"u}ller, H.:
  Vqa-med: Overview of the medical visual question answering task at imageclef
  2019. In: CLEF. 9-12 September 2019 (2019)

\bibitem{chen2023towards}
Chen, Z., Diao, S., Wang, B., Li, G., Wan, X.: Towards unifying medical
  vision-and-language pre-training via soft prompts. arXiv preprint
  arXiv:2302.08958  (2023)

\bibitem{do2021multiple}
Do, T., Nguyen, B.X., Tjiputra, E., Tran, M., Tran, Q.D., Nguyen, A.: Multiple
  meta-model quantifying for medical visual question answering. In: MICCAI. pp.
  64--74. Springer (2021)

\bibitem{dou2022empirical}
Dou, Z.Y., Xu, Y., Gan, Z., Wang, J., Wang, S., Wang, L., Zhu, C., Zhang, P.,
  Yuan, L., Peng, N., et~al.: An empirical study of training end-to-end
  vision-and-language transformers. In: CVPR. pp. 18166--18176 (2022)

\bibitem{eslami2021does}
Eslami, S., de~Melo, G., Meinel, C.: Does clip benefit visual question
  answering in the medical domain as much as it does in the general domain?
  arXiv preprint arXiv:2112.13906  (2021)

\bibitem{finn2017model}
Finn, C., Abbeel, P., Levine, S.: Model-agnostic meta-learning for fast
  adaptation of deep networks. In: ICML. pp. 1126--1135. PMLR (2017)

\bibitem{gong2021cross}
Gong, H., Chen, G., Liu, S., Yu, Y., Li, G.: Cross-modal self-attention with
  multi-task pre-training for medical visual question answering. In: ACM ICMR.
  pp. 456--460 (2021)

\bibitem{gong2022vqamix}
Gong, H., Chen, G., Mao, M., Li, Z., Li, G.: Vqamix: Conditional triplet mixup
  for medical visual question answering. IEEE Trans Med Imaging
  \textbf{41}(11),  3332--3343 (2022)

\bibitem{he2020pathological}
He, X., Cai, Z., Wei, W., Zhang, Y., Mou, L., Xing, E., Xie, P.: Pathological
  visual question answering. arXiv preprint arXiv:2010.12435  (2020)

\bibitem{huang2022ovqa}
Huang, Y., Wang, X., Liu, F., Huang, G.: Ovqa: a clinically generated visual
  question answering dataset. In: ACM SIGIR. pp. 2924--2938 (2022)

\bibitem{huang2023effective}
Huang, Y., Wang, X., Su, J.: An effective pre-trained visual encoder for
  medical visual question answering. In: ADMA. pp. 466--481. Springer (2023)

\bibitem{khare2021mmbert}
Khare, Y., Bagal, V., Mathew, M., Devi, A., Priyakumar, U.D., Jawahar, C.:
  Mmbert: Multimodal bert pretraining for improved medical vqa. In: ISBI. pp.
  1033--1036. IEEE (2021)

\bibitem{lau2018dataset}
Lau, J.J., Gayen, S., Ben~Abacha, A., Demner-Fushman, D.: A dataset of
  clinically generated visual questions and answers about radiology images.
  Scientific data  \textbf{5}(1),  1--10 (2018)

\bibitem{liu2021slake}
Liu, B., Zhan, L.M., Xu, L., Ma, L., Yang, Y., Wu, X.M.: Slake: A
  semantically-labeled knowledge-enhanced dataset for medical visual question
  answering. In: ISBI. pp. 1650--1654. IEEE (2021)

\bibitem{masci2011stacked}
Masci, J., Meier, U., Cire{\c{s}}an, D., Schmidhuber, J.: Stacked convolutional
  auto-encoders for hierarchical feature extraction. In: ICANN. pp. 52--59.
  Springer (2011)

\bibitem{nguyen2019overcoming}
Nguyen, B.D., Do, T.T., Nguyen, B.X., Do, T., Tjiputra, E., Tran, Q.D.:
  Overcoming data limitation in medical visual question answering. In: MICCAI.
  pp. 522--530. Springer (2019)

\bibitem{pelka2018radiology}
Pelka, O., Koitka, S., R{\"u}ckert, J., Nensa, F., Friedrich, C.M.: Radiology
  objects in context (roco): a multimodal image dataset. In: LABELS. pp.
  180--189. Springer (2018)

\bibitem{Srivastava_Murali_Dubey_Mukherjee_2021}
Srivastava, Y., Murali, V., Dubey, S.R., Mukherjee, S.: Visual Question
  Answering using Deep Learning: A Survey and Performance Analysis, p. 75–86
  (Jan 2021). \doi{10.1007/978-981-16-1092-9\_7},
  \url{http://dx.doi.org/10.1007/978-981-16-1092-9_7}

\bibitem{Wu_Teney_Wang_Shen_Dick_Hengel_2016}
Wu, Q., Teney, D., Wang, P., Shen, C., Dick, A., Hengel, A.: Visual question
  answering: A survey of methods and datasets. Cornell University -
  arXiv,Cornell University - arXiv  (Jul 2016)

\bibitem{yang2022vision}
Yang, J., Duan, J., Tran, S., Xu, Y., Chanda, S., Chen, L., Zeng, B., Chilimbi,
  T., Huang, J.: Vision-language pre-training with triple contrastive learning.
  In: CVPR. pp. 15671--15680 (2022)

\bibitem{yasunaga-etal-2022-linkbert}
Yasunaga, M., Leskovec, J., Liang, P.: {L}ink{BERT}: Pretraining language
  models with document links. In: ACL. pp. 8003--8016 (2022)

\bibitem{zhan2020medical}
Zhan, L.M., Liu, B., Fan, L., Chen, J., Wu, X.M.: Medical visual question
  answering via conditional reasoning. In: ACM MM. pp. 2345--2354 (2020)

\bibitem{zhu2023minigpt}
Zhu, D., Chen, J., Shen, X., Li, X., Elhoseiny, M.: Minigpt-4: Enhancing
  vision-language understanding with advanced large language models. arXiv
  preprint arXiv:2304.10592  (2023)

\end{thebibliography}
\end{document}